\documentclass[conference,a4paper]{APSIPA2025}
\usepackage{amsmath}
\usepackage{graphicx}
\usepackage{multirow}
\usepackage{threeparttable}
\usepackage[backend=biber,style=ieee,]{biblatex}
\usepackage{amssymb,amsfonts}
\usepackage{algorithmic}
\usepackage{textcomp}
\usepackage{xcolor}
\usepackage{subcaption}
\usepackage{caption}
\DeclareCaptionJustification{justified}{\justifying}
\usepackage[raggedrightboxes]{ragged2e}
\usepackage{algorithm}
\usepackage{algorithmic}
\usepackage{float}
\usepackage{tabularx}
\usepackage{float}
\usepackage{array}
\usepackage{moresize}
\usepackage{geometry}
\usepackage{fancyhdr}
\usepackage[normalem]{ulem}
\useunder{\uline}{\ul}{}

\fancypagestyle{firststyle}{
  \fancyhf{}
  \fancyhead[C]{}
}

\begin{document}

\title{A Rate-Quality Model for Learned Video Coding}

\author{
\authorblockN{
Sang NguyenQuang\authorrefmark{1}, Cheng-Wei Chen\authorrefmark{1}, Xiem HoangVan\authorrefmark{2} and
Wen-Hsiao Peng\authorrefmark{1}
}

\authorblockA{
\authorrefmark{1}
National Yang Ming Chiao Tung University, Taiwan }

\authorblockA{
\authorrefmark{2}
VNU University of Engineering and Technology, Vietnam 
}
}

\maketitle
\thispagestyle{firststyle}

\begin{abstract}
Learned video coding (LVC) has recently achieved superior coding performance. However, there is a lack of research on the rate-quality (R-Q) model which is important in real-time LVC applications. In this paper, we propose a parametric model to characterize the R–Q relationship in LVC systems. In the proposed model, a neural network, termed RQNet is introduced to characterize the relationship between bitrate and quality level according to video content and coding context. The predicted (R,Q) results are further integrated with those from previously coded frames using the least-squares method to determine the parameters of our R-Q model on-the-fly. Compared to the conventional approaches, our method accurately estimates the R-Q relationship, enabling the online adaptation of model parameters to enhance both flexibility and precision. Experimental results show that our R-Q model achieves significantly smaller bitrate deviations than the baseline methods on commonly used datasets with minimal additional complexity.

%Our framework is applied to solve the rate control problem in LVC. 
% Rate control remains a critical challenge in learning-based video compression. Most existing rate control approaches suffer from inaccurate rate-quality (R-Q) models and insufficient adaptation to video content. To address this issue, we develop an efficient R-Q model for learned video compression (RQ-LVC). 
%. from  adapts to various video sequences. On commonly used datasets, our approach achieves a comparable bitrate error as the existing method 
% Rate control for learning-based video compression is a classic problem. However, an essential challenge in this scheme is the accuracy of modeling and controlling the rate-quality (R-Q) relationship, which directly impacts the efficiency of bitrate allocation and overall compression performance. To address this issue, we propose an efficient R-Q model for learning-based video compression. Firstly, we formulate a mathematical function to efficiently represent the R-Q behaviors. Then, we introduce a parameter estimation method that dynamically adjusts the model for each coding frame. Finally, we employ a deep neural network to estimate the parameters governing this relationship by leveraging both the content and contextual information of the current frame, as well as encoding statistics from preceding frames. 
% but reduces almost 96\% kMAC/pixel in terms of Full fitting set and 33\% kMAC/pixel in terms of Sub fitting set.

\end{abstract}
\section{Introduction}
\label{sec:intro}
Learned video coding has been considered as a promising coding solution for a wide range of video transmission and storage. Similar to traditional video coding standards, rate control plays a crucial role in deploying LVC for practical video applications. An effective rate control system must deliver high rate-distortion performance while minimizing deviation between actual and target bitrates. The encoder's efficiency depends on bitrate allocation across frames, while bitrate deviation is determined by encoding parameters. To achieve this, a well-designed model that accurately captures the relationship between bitrate and coding parameters (i.e., quantization parameter or quality level) is indispensable.

In learned video coding~\cite{canfvc,maskcrt,dcvcdc,dcvcfm}, accurately estimating the rate-quality (R-Q) relationship poses a significant challenge. Unlike traditional codecs~\cite{h264,hevc,overview_vvc} with well-calibrated analytical models, learned video codecs exhibit distinct R-Q characteristics heavily influenced by model architectures, training protocols, and optimization strategies.
% In addition, image and video coding methods have incorporated rate control, a crucial technique widely used in practical applications. To accurately predict encoding parameters for achieving the target bitrate, the relationship between rate (R), distortion (D), and slope of the R-D curve $(\lambda)$ has been widely modeled, leading to significant progress in its applications over the past few years. 
% However, modeling the rate-quality relationship remains significantly unexplored.
% However, despite recent state-of-the-art learned video coding~\cite{dvclu, fvc, mlvc, elfvc, dcvcdc, dcvcfm}, especially transformer-based approaches~\cite{maskcrt}, the relationship between bitrate and distortion (or quality) remains underexplored. 
% Unlike conventional codecs, where R-D behavior follows well-defined analytical models, learned codecs~\cite{dvclu,dcvc,fvc,canfvc,maskcrt} exhibit completely different rate-quality characteristics influenced by several factors such as model architecture, training protocols, and optimization strategies. The accurate estimation of the relationship between rate and quality remains a critical challenge for learned video coding.
% studied for many years. However, for learned video coding~\cite{dvclu, fvc, mlvc, elfvc, dcvcdc, dcvcfm}, particularly transformer-based video coding~\cite{maskcrt}, there are very few specialized studies that have identified this relationship. For these codecs, the relationship between rate-distortion or rate-quality depends on the characteristics of each codec, such as the backbone, training procedure, etc.

\begin{figure}[t]
\centering
\includegraphics[width=0.9\linewidth]{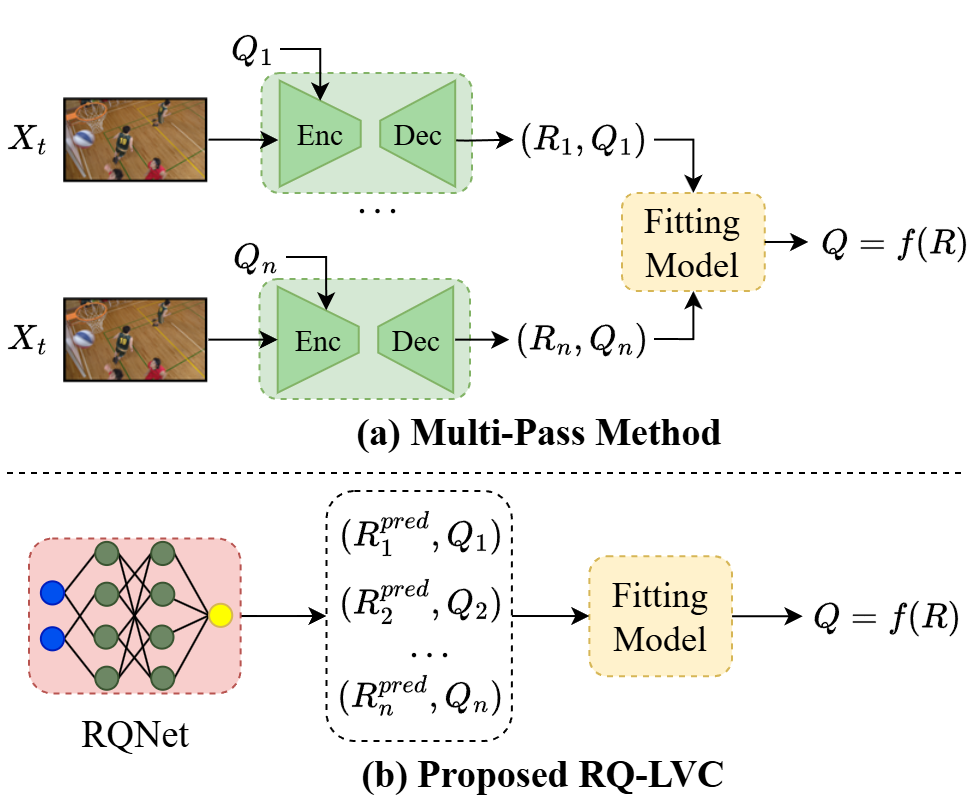}
    %\vspace{-3em}
\caption{Comparison of the conventional method and our RQ-LVC in predicting rate-quality relationship.}
\label{fig:teaser}
\vspace{-1.7em}
\end{figure}

% The flexibility of rate control in image and video coding is constrained by the predefined discrete $\lambda$ values, which limit its adaptability to diverse video content and coding scenarios. Prior research has explored various approaches to model the relationship between R, D, and $\lambda$. 
In recent years, capturing R-Q relationships in image and video coding has emerged as an important research topic. Jia~\emph{et al.}~\cite{rd_modeling_lic} utilize exponential and logarithmic functions to characterize the relationship among rate (R), distortion (D), and the Lagrange multiplier ($\lambda$), which indicates the slope
of the R-D curve. Xue~\emph{et al.}~\cite{LambdaDomainRC_LIC} propose an exponential function with an additional bias term to model the R-$\lambda$ relationship and enhance the rate-fitting accuracy. Similar approaches have been applied to video coding~\cite{LiLiLambda}, where the R-$\lambda$ and D-$\lambda$ curves are fitted using exponential models, and $\lambda$ values are updated dynamically based on encoding results of previous frames. Inspired by~\cite{LiLiLambda}, Li~\emph{et al.}~\cite{rate_control_tencent} propose a learned R-D-$\lambda$ model and parameter updating mechanism for learned video coding. Jia~\emph{et al.}~\cite{RC_ICASSP} explore the relationship between bitrate and quantization parameter, while Chen~\emph{et al.}~\cite{SparseToDense} introduce the scaling factor as a hyperparameter in the R-D model for learned video codecs. Additionally, Gu~\emph{et al.}~\cite{rate_fitting} design two neural network models to predict (R,$\lambda$) and (D,$\lambda$) points, leveraging them to model the relationship between R and D. In contrast, Zhang~\emph{et al.}~\cite{neural_rate_control} propose a fully neural network-based rate control system including a rate allocation model and rate implementation network to perform the rate-parameter mapping. Although these methods have proven effective, they remain fundamentally constrained by predefined $\lambda$ values, limiting their adaptability for rate control in learned video coding. Moreover, most methods either update model parameters iteratively by adjusting parameters based on rate-distortion statistics collected from previously encoded frames or use neural networks to predict coding parameters. These approaches operate separately and do not take full advantage of available information.
% propose modeling the relationship between rate, distortion, and scaling factors, applying it to bitrate control in scale-adaptive video coding. However, the limitation of these methods lies in the use of a predefined set of discrete $\lambda$ values, which restricts the flexibility of the encoding process.

In this paper, we introduce a rate-quality (R-Q) model for learned video coding to estimate the R-Q relationship. In our approach, termed RQ-LVC, the quality level acts as a coding parameter to regulate the variable-rate video coding process. Compared with conventional approaches that independently model the R-$\lambda$ or D-$\lambda$ relationships, our method directly models the link between bitrate and quality, enabling precise rate control for learning-based video codecs. Furthermore, our proposed framework dynamically updates the R-Q model on-the-fly in a batch manner, allowing online adaptation by using information from previously coded frames and predictions from a deep neural network.
% As illustrated in Fig~\ref{fig:teaser}, the proposed method models R-Q relationship efficiently. 

Our contributions in this work are best illustrated in Fig.~\ref{fig:teaser}. As depicted, multi-pass approaches collect (R,Q) points by encoding a video frame multiple times with different quality levels to fit the R-Q relationship, which is highly time-consuming. In contrast, our proposed RQ-LVC efficiently models the R-Q relationship with low complexity. By leveraging RQNet, our approach significantly reduces encoding complexity while ensuring accurate rate control, offering a more computationally efficient and scalable solution. In Table~\ref{tab:compare_method}, we compare the proposed method with several prior works in terms of the rate control model, model parameters updating mechanism, and application scenario.
\begin{table}[t]
\centering
% \scriptsize
\caption{Our method vs. prior works}
\fontsize{6.8}{9}\selectfont
\begin{tabular}{ccccc}
\hline
\textbf{Method}    & \begin{tabular}[c]{@{}c@{}}\textbf{Rate Control}\\ \textbf{Model}\end{tabular} & \begin{tabular}[c]{@{}c@{}}\textbf{Updating}\\ \textbf{Mechanism}\end{tabular} & \textbf{Application}                                                 \\ \hline
TIP'14~\cite{LiLiLambda}    		& $\lambda =\alpha R^{\beta }$		& Iterative Mode		& Traditional Codec \\ \hline
% ICASSP'24~\cite{RC_ICASSP} 		& \begin{tabular}[c]{@{}c@{}}$R=C\times Q^{-K}$\\ $\Leftrightarrow Q=\left ( \frac{C}{R} \right )^{1/K}$\end{tabular}		& Iterative Mode		& Learned Codec     \\ \hline
ICASSP'24~\cite{RC_ICASSP} 		& $R=C\times Q^{-K}$		& Iterative Mode		& Learned Codec  \\ \hline
ICLR'24~\cite{neural_rate_control}  		& Neural Network		& Neural Network		& Learned Codec     \\ \hline
TCSVT'24~\cite{SparseToDense}  		& $R=\alpha\times r^\beta$		& Iterative Mode		& Learned Codec     \\ \hline
DCC'25~\cite{rate_fitting}   		& $R=\alpha\times\lambda^\beta$		& Neural Network		& Learned Codec     \\ \hline
Our RQ-LVC      		& $Q = \alpha \times \operatorname{ln}(R) + \beta$		& Batch Mode		& Learned Codec     \\ \hline
\end{tabular}
\begin{flushleft}
*R: bitrate; Q: quality level; $\lambda$: Lagrange multiplier; $\alpha$, $\beta$, $C$, $K$: model parameter; $r$: rescale ratio.
\end{flushleft}
\label{tab:compare_method}
\end{table}
%To enhance prediction accuracy across various video content, the R-Q estimation function is iteratively updated on-the-fly during the encoding process, ensuring reliable bitrate estimation under diverse encoding conditions.

% we present a method for training a variable rate model for video coding, enabling the encoder to cover a wide range of bitrates and quality levels. The approach focuses on modeling the rate-quality (R-Q) relationship and dynamically updating its parameters for accurate rate control in learning-based video codecs. Unlike independent modeling of R-D or R-$\lambda$ relationships, we directly model the bitrate-quality link. By leveraging neural networks, the method estimates and updates these parameters in real-time based on encoding results from previous frames. The process starts by collecting data from frames encoded at various quality levels and fitting a parameterized function to describe the R-Q relationship, which is continuously refined using a weighted curve fitting approach to ensure accurate bitrate predictions even as video content changes. 

% The rest of this paper is organized as follows. Section II presents the work flows and architecture of our proposed RQ-LVC. Afterwards, Section 4 discusses the model assessment and the coding efficiency. Finally, Section 5 gives some conclusions and outlines future works. 

\section{Proposed Method}
\begin{figure}[t]
\centering
\includegraphics[width=\linewidth]{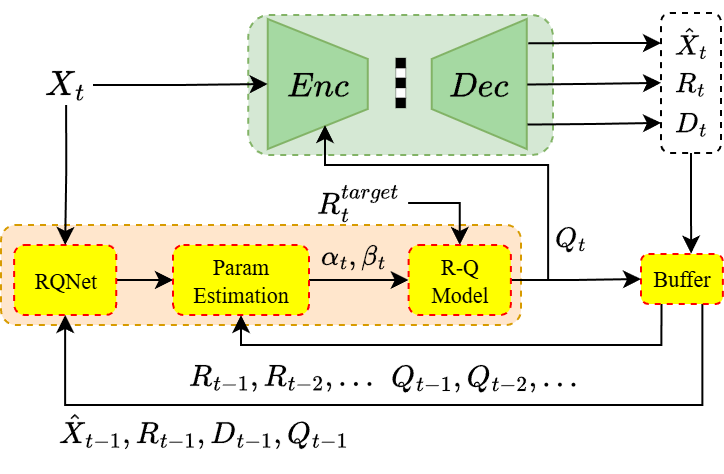}
    %\vspace{-3em}
\caption{System overview of our RQ-LVC framework with the RQNet, parameter estimation, and R-Q model. 
% We do a statistical analysis to model the relationship between Rate and Quality (R-Q). 
RQNet extracts spatiotemporal information to predict (R,Q) points, which are combined with those from previously coded frames to estimate the parametric R-Q model. Then, the predicted $Q_t$ is estimated based on the target bitrate $R^{target}_{t}$ to regularize the encoding of an input video frame.}
\label{fig:proposed}
\vspace{-1.5em}
\end{figure}
\label{sec:method}
Fig.~\ref{fig:proposed} illustrates the overall architecture of the proposed rate-quality model for learned video coding (RQ-LVC). RQ-LVC encodes each input frame $X_t$ at variable quality level $Q_t$ under a rate constraint $R^{target}_t$ by modeling the rate-quality (R-Q) relationship. Specifically, our proposed plug-in module, RQNet, predicts (R,Q) points based on the coding context--such as rate and distortion of reference frames--and content of $X_t$. These predictions serve as prior knowledge, which is further refined using the observed (R,Q) points collected from previously coded frames to update the R-Q model. The updated R-Q model determines the appropriate quality level $Q_t$ to satisfy $R^{target}_{t}$.

%, a latest learned transformer-based variable-rate P-frame video coding% .
% (termed RQ-LVC) architecture, which is a variable-rate video coding framework built on top of the most recent P-frame learning video encoder \cite{maskcrt}. Our RQ-LVC is trained to be able to encode an input video at multiple quality levels. We model the rate-quality relationship (RQ model) through 
% In order to perform rate control, RQNet is used to predict a set of $(R, Q)$ points of the coding frame, then combine them with the (R, Q) points coming from the previous coded frames to estimate the RQ model parameters. This model is then used to derive the quality level $Q$ based on the target bitrate $R^{target}$.
\subsection{Learning a Variable-Rate Neural Video Codec}
\label{ssec:variable_model}
Our framework features a variable-rate neural video codec that adapts bitrate and quality based on quality level $Q$. Instead of training separate models for different rates, we develop a variable-rate version using MaskCRT~\cite{maskcrt} as the base codec.
% Our framework features a MaskCRT-based~\cite{maskcrt} variable-rate

To train this variable-rate version of MaskCRT~\cite{maskcrt}, we adopt the standard training loss function for learned video coding, namely, $Loss_{RD}=R+\lambda D$, 
where $R$, $D$, and $\lambda$ represent the bitrate, distortion, and Lagrange multiplier, respectively. To support wide bitrate ranges with a single model, $\lambda$ is randomly sampled from a predefined range during each training iteration. Following~\cite{dcvcfm}, we relate $\lambda$ to a (continuous) quality parameter $Q$ within the interval $[0,Q_{num}-1]$: 
\begin{equation}
\label{equ:lambda}
\lambda = \exp\left\{\ln\lambda_{min} + \frac{Q}{Q_{num}-1}  (\ln \lambda_{max} - \ln \lambda_{min})\right\}, 
% \lambda =\exp \left\{\operatorname{ln}\left(\lambda_{min} +\frac{Q_t}{Q_{num}-1}\cdot (\lambda _{max}-\lambda_{min})\right)\right\},
\end{equation}
where $\lambda_{min},\lambda_{max}$ define the lowest and highest Lagrange multipliers, respectively. To sample a $\lambda$ between $\lambda_{min}$ and $\lambda_{max}$, we simply choose a $Q$ from $[0,Q_{num}-1]$. $Q_{num}=64$ in our current implementation.

%This logarithmic interpolation ensures a relationship between quality and bitrate that aligns with conventional video codec. Consequently, the model learns to encode video at various quality parameters in frame level, enhancing its adaptability and effectiveness in rate fitting tasks.
%, so we simply adopt a common variable-rate approach \cite{dcvcfm}, where $\lambda$ values are continuously sampled within a predefined range at each training step. Specifically, we randomly  assign a quality level $q_t$ from a range $[0, q_{num}-1]$ and interpolate the corresponding $\lambda$ value with the following formula:

%The variable quality model is able to encode the video sequences with different quality levels of coding via adjusting the quality parameter $Q$ directly.
% By adjusting the value of $\lambda$, the encoder can regulate the level of coding, allowing it to encode the input video at different quality levels.
% The optimization objective of a video codec is typically defined as $Loss_{RD}=R+\lambda D$, 

\subsection{Parametric Rate-Quality (R-Q) Models}
% \subsection{Rate-Quality Modeling}
\label{ssec:rq_modeling}

\begin{figure}[t!]
    \begin{center}
    \begin{subfigure}{\linewidth}
        \centering
        \includegraphics[width=0.8\linewidth]{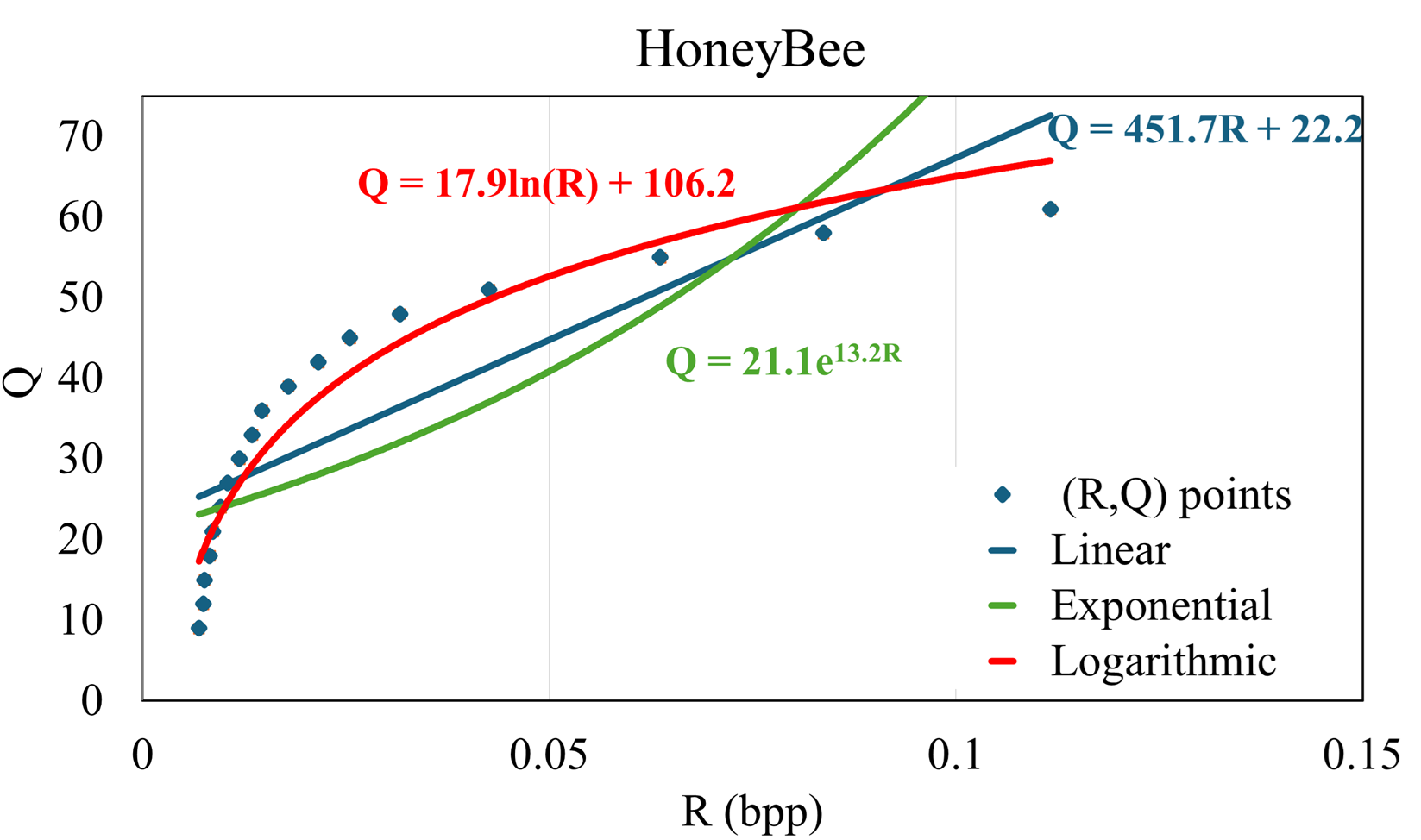}
    \end{subfigure}
    \begin{subfigure}{\linewidth}
    \vspace{0.5em}
        \centering
        \includegraphics[width=0.8\linewidth]{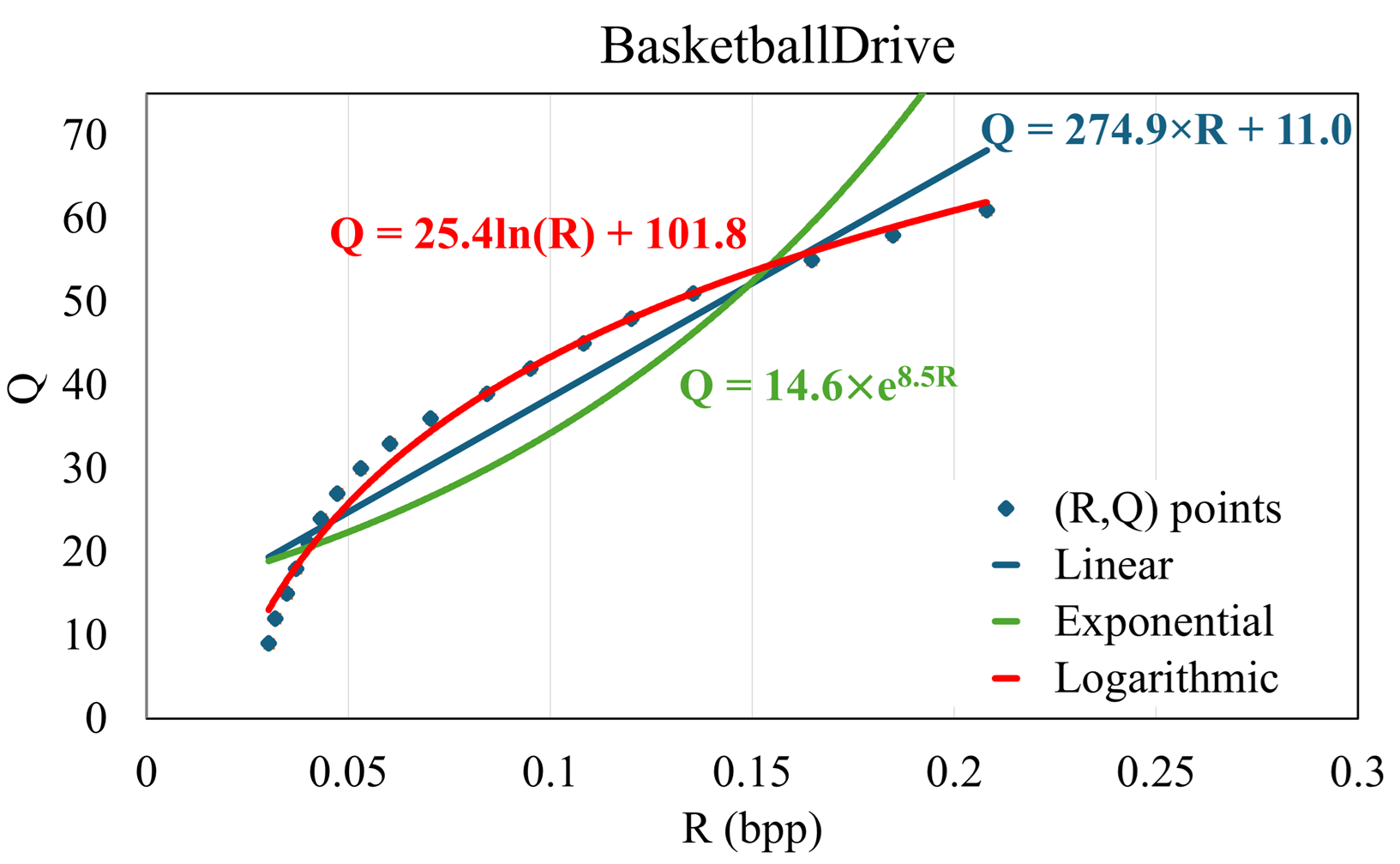}
        \label{fig:honey_scale}
        % \caption{ }
    \end{subfigure}
    %\vspace{-3em}
    \caption{R-Q models for HoneyBee and BasketballDrive sequences.}
    \label{fig:curve_fitting}
    \end{center}
    \vspace{-.5em}
\end{figure}

We adopt a parametric approach to construct an R-Q model, enabling the prediction of $Q$ for a given target bitrate $R^{target}$. To analyze the relationship between the bitrate $R$ and quality level $Q$, we collect frame-level (R,Q) statistics from the UVG~\cite{uvg} and HEVC Class B~\cite{hevcctc} datasets. This is achieved by encoding some sequences from these datasets using various quality levels $Q$ and collecting the resulting bitrates $R$. Three parametric models--linear, exponential, and logarithmic models--are employed to fit these frame-level (R,Q) data points:
    \begin{equation}
    \label{equ:linear}
    Q = \alpha \times R + \beta,
    \vspace*{-.1cm}
    \end{equation}
    \begin{equation}
    \label{equ:exp}
    Q = \alpha \times e^{\beta },
    \vspace*{-.1cm}
    \end{equation}
    \begin{equation}
    \label{equ:log}
    Q = \alpha \times \operatorname{ln}(R) + \beta,
    \vspace*{-.1cm}
    \end{equation}
where $\alpha,\beta$ are obtained by the least-squares method. In this work, the model achieving the highest $R^{2}$ score is deemed the best among the three. As shown in Table~\ref{tab:r_squared}, the logarithmic model consistently outperforms the linear and exponential models. Furthermore, Fig.~\ref{fig:curve_fitting} confirms that the logarithmic model effectively captures the relationship between bitrate ($R$) and quality level ($Q$). Thus, we choose the logarithmic model as our parametric solution for modeling the R-Q relationship.
% All videos have $R^{2}$ values greater than 0.9 when applying the logarithmic model, while the $R^{2}$ values of the linear model and exponential model functions are significantly lower. Therefore, in this paper, we use the logarithmic function to model the R-Q relationship.

% \subsection{Rate-Quality Modeling}
\subsection{Online Estimation of Model Parameters}
The coefficients $\alpha$ and $\beta$ of the R-Q model are highly dependent on both video content and the backbone codec (e.g., MaskCRT). A straightforward approach to estimate their values involves encoding an input video multiple times using various quality levels and fitting $\alpha,\beta$ to the resulting frame-level (R,Q) data points. However, this approach is computationally expensive and impractical. Drawing inspiration from Bayesian inference~\cite{bayesian}, we learn a neural network, termed RQNet, to predict a number of (R,Q) points based on video content and contextual information. Because RQNet is trained offline on a large dataset, these network-predicted (R,Q) points provide the prior knowledge of the R-Q relationship for the current coding context. To address the amortized inference issue (i.e., the RQNet predictions are not optimal for individual videos), we further refine this prior knowledge by incorporating the (R,Q) points collected from previously coded frames.
These (R,Q) points are then combined in a least-squares framework to dynamically update $\alpha$ and $\beta$ at inference time.

%"After collecting these (R, Q) points, $\alpha,\beta$ can be obtained by performing curve fitting. However, we observed that (R, Q) points predicted by RQNet and those come from actuall encoding process have different impact. Similarly, data points of more recent frames carry greater relevance compared to earlier frame. 
%All these (R,Q) points are utilized together in a weighted least-squares framework to update $\alpha,\beta$ on the fly. 

%In this approach, the bitrate required to encode video frames at specific quality levels can be estimated by a neural network, called RQNet, without performing the entire encoding process. Video content and contextual information are used to approximate (R, Q) points precisely.

% we employ neural network to estimate the bitrate required to encode the current frame at specific quality levels, based on content and contextual information of the frame, to approximate (R, Q) points without performing the full encoding process. The encoding results of previously encoded frames are also stored as (R, Q) points, serving as prior knowledge. Curve fitting is then applied to these points to obtain the model parameters. Our method effectively updating the parameter estimates as new encoding results become available, analogous to the posterior distribution in Bayesian inference.

% \begin{figure}[t]
%     %\vspace{-1em}
%         \centering
%         \includegraphics[width=0.8\linewidth]{figures/BasketballDrive.png}
%     %\vspace{-3em}
%     \caption{R-Q fitting models for BasketballDrive.}
%     \label{fig:curve_fitting}
%     \vspace{-0.3cm}
% \end{figure}
\begin{table}[t]
\caption{$R^{2}$ scores for various parametric models.}
% \small
% \footnotesize
\scriptsize
\setlength{\tabcolsep}{5pt}
\centering
\begin{tabular}{cccc}
\hline
\multirow{2}{*}{{Sequence}} & \multicolumn{3}{c}{{Fitting Model}}                  \\ \cline{2-4} 
                                   & {Linear} & {Exponential} & {Logarithmic} \\ \hline
Kimono1                            & 0.943           & 0.806                & 0.997              \\
BasketballDrive                    & 0.898           & 0.767                & 0.988              \\
ReadySteadyGo                      & 0.925           & 0.781                & 0.995              \\
Jockey                             & 0.801           & 0.680                & 0.964              \\
HoneyBee                           & 0.684           & 0.547                & 0.929              \\
Beauty                             & 0.810           & 0.713                & 0.948              \\
\textit{Average}                            & \textit{0.844}           & \textit{0.716}               & \textit{0.970}             \\\hline
\end{tabular}
\label{tab:r_squared}
\vspace{-0.3cm}
\end{table}

\subsubsection{RQNet}
\label{sssec:param_init}
\begin{figure}[t]
\centering
    \captionsetup[subfigure]{justification=centering}
    \begin{subfigure}{0.45\textwidth}
        \centering
        \includegraphics[width=0.85\linewidth]{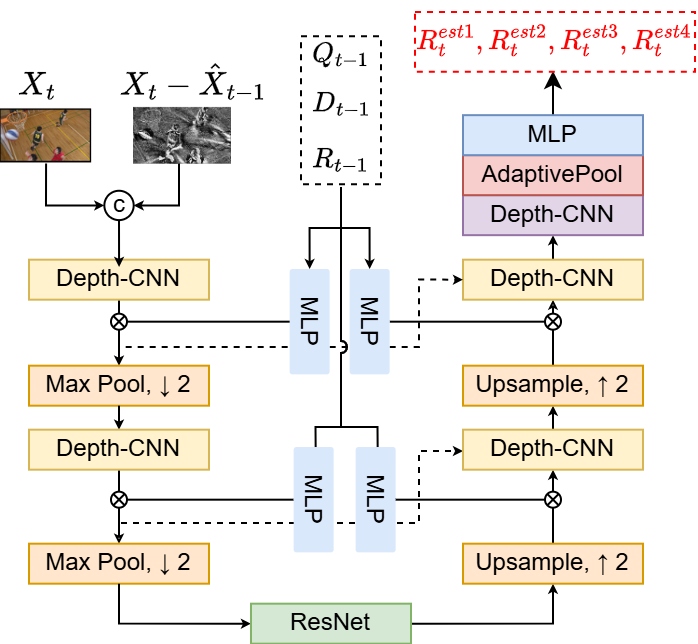}
        % \label{fig:rq_model}
        %\vspace{-3em}
    \end{subfigure}
% \includegraphics[width=0.65\linewidth]{figures/Rate_Quality_Predictor.png}
    %\vspace{-3em}
    % \hfill
    % \hspace{6mm}
    
    % \begin{subfigure}{0.5\textwidth}
    %     \centering
    %     \includegraphics[width=0.55\linewidth]{figures/Depth_CNN.png}
    %     \caption{The Depth-CNN}
    %     % \label{fig:depth_cnn}
    % \end{subfigure}
\caption{The architecture of the proposed RQNet. Depth-wise separable convolution (Depth-CNN)~\cite{depthCNN} and MLP layers are used to extract spatiotemporal information.}
\label{fig:rq_model}
\vspace{-1.7em}
\end{figure}
Fig.~\ref{fig:rq_model} depicts the U-Net architecture of RQNet. The inputs consist of the current coding frame $X_t$, frame difference $X_{t}-\hat{X}_{t-1}$ between $X_t$ and the previously decoded frame $\hat{X}_{t-1}$, as well as the rate-distortion statistics ${R}_{t-1}$, ${D}_{t-1}$, $Q_{t-1}$ from the previous frame. The model outputs four bitrates corresponding to the predefined quality levels of 10, 17, 43 and 60, respectively. Instead of predicting the bitrate for every possible quality level, we simplify the task to focus on only four bitrates for these fixed quality levels. Consequently, RQNet provides four (R,Q) points that are both content- and context-dependent.

%By performing deep downscaling and upscaling in feature extraction, RQNet captures richer representations, enabling it to predict the corresponding bitrate precisely. 
% The details of our RQNet are presented in Fig.~\ref{fig:rq_model}. The input consists of the current frame, the reconstructed previous frame, and historical coding parameters, including rate, distortion, and quantization values of the previous frame. The content information is extracted from current and reference frame through depthwise CNNs with max-pooling, while past coding parameters are embedded using MLPs. The fused features are further enhanced by a ResNet and then refined through upsampling, additional CNN layers, and adaptive pooling, with a final MLP producing the predicted bitrate.
% \textcolor{red}{Write about model design and training strategy}
% Our proposed RQNet establishes a mapping between the target bitrate $R_{target}$ and the quality level $Q_{pred}$ for variable-rate models, as shown in Fig~\ref{fig:rq_model}. Inspired by the approach in \cite{neural_rate_control}, we model the relationship between the target bitrate and the coding parameter as a regression task, where the prediction is influenced by the content properties of the current frame and the encoding outcomes of the previous frame. Since variations in content between consecutive frames significantly influence the coding bitrate, we further incorporate state signals from the previous encoding frame to enhance the accuracy of rate-quality prediction.
To train RQNet, video frames are encoded at predefined quality levels to collect the actual bitrates, denoted as ${R}^{enc}_t$. The model then predicts the estimated bitrates ${R}^{pred}_t$ for the four quality levels. The training objective is to minimize the mean absolute deviation between $R_t$ and ${R}^{pred}_t$:
\begin{equation}
\label{equ:loss}
L = \frac{1}{4} \sum^4_{n=1}  \left | {R^{enc(n)}_{t}-{{{R}^{pred(n)}_{t}}}} \right|.
\end{equation}
\subsubsection{Least-Squares for Parameter Estimation}
\label{sssec:param_update} 
There are two sets of (R,Q) points. One is derived from the output of RQNet and the other is collected from the actual encoding of previous frames. 
The (R,Q) points from RQNet serve as prior knowledge while those sampled from coded frames are additional observations.

Consider the data points $\{R_t^{\text{pred}(n)},Q_t^{\text{pred}(n)}\}_{n=1}^4$ from RQNet and $\{R_j^{enc},Q_j^{enc}\}_{j<t}$ collected from the coded frames up to the time index $t$ within a group-of-pictures (GOP); $\alpha_t$ and $\beta_t$ are estimated by minimizing the sum of squared errors:
\begin{align}
\label{equ:curve_fitting}
    \arg\min_{\alpha_t, \beta_t} \quad & \sum_{n=1}^{4}  \left( Q_t^{pred(n)} - f\left( R_t^{pred(n)}; \alpha_t, \beta_t \right) \right)^2 \notag \\
    & + \sum_{j=1}^{t-1} \left( Q_j^{enc} - f\left( R_j^{enc}; \alpha_t, \beta_t \right) \right)^2,
\end{align}
where $f(\cdot)$ represents the logarithmic R-Q function in Eq.~\eqref{equ:log}.

\section{Experimental Results}
\label{sec:results}
\subsection{Experimental Settings}
\label{ssec:exp_setting}
\textbf{Datasets:} We use Vimeo-90K~\cite{vimeo} to train RQNet and the variable-rate MaskCRT. We randomly crop 256$\times$256 patches for training. The test datasets are HEVC Class B~\cite{hevcctc} and UVG~\cite{uvg}.

\textbf{Implementations:} We evaluate our proposed method with three variants: RQ-LVC, RQ-LVC w/ RQNet Only, and RQ-LVC w/o RQNet. Our RQ-LVC combines (R,Q) points collected from both RQNet and previously coded frames to estimate the parameters of the R-Q model. In contrast, RQ-LVC w/ RQNet relies solely on RQNet predictions, while RQ-LVC w/o RQNet uses only previously coded frames without utilizing any predictions from RQNet.

\textbf{Evaluation~Methodologies:}In our evaluation, we incorporate proposed RQ-LVC into a rate allocation algorithm, as in~\cite{LiLiLambda, RC_ICASSP}. First, we determine the average target bitrates per frame $R_s$ using MaskCRT~\cite{maskcrt} with a fixed quality level $Q\in \left\{ 10, 25, 40, 55\right\}$. Based on the given $R_s$, we compute the target bitrate $R_{mg}$ for a miniGOP, which includes a set of $N_m$ consecutive frames $\left\{X_{i},X_{i+1},...,X_{t},...,X_{i+N_{m}-1}\right\}$:
\begin{equation}
\label{equ:minigop_allocate}
R_{mg}=\frac{R_{s}\times\left ( N_{coded}+SW \right )-\hat{R_{s}}}{SW} \times N_{m},
% \vspace{-0.3cm}
\end{equation}
where $N_{coded}$ is the number of encoded frames, $\hat{R_{s}}$ is the total bitrate already consumed, and $SW$ refers to the sliding window size, which is set to 40 in our implementation. Afterward, the target bitrate $R_t$ (in bits) for frame $X_t$ is given by:
\begin{equation}
\label{equ:frame_allocate}
R_{t}=\frac{R_{mg}-\hat{R}_{mg}}{\sum_{j=t}^{i+N_{m}-1}}\times w_{t},
% \vspace{-0.3cm}
\end{equation}
where $\hat{R}_{mg}$ is the bitrate consumed by previously coded frames within the current miniGOP, and $w_t$ denotes the rate allocation weight for frame $X_t$. In this work, we set the number $N_m$ of frames in a miniGOP to 4 with empirically chosen weights $\left\{1.9,1.6,1.3,1.0 \right\}$.

The experiments are conducted by encoding the first 96 frames of each video sequence. We pre-encode the video sequences using a single quality level across all frames, and then use the resulting bitrate as the target bitrate for RQ-LVC. We measure $\Delta R^{RC}$ to assess RQ-LVC rate control accuracy:
\begin{equation}
\label{equ:delta_R_RC}
\Delta R^{RC} = \left | \frac{R^{target}-R^{enc}}{R^{target}} \right |\times100\%,
\end{equation}
where $R^{target}$ and $R^{enc}$ denote the target bitrate and the actual bitrate resulting from encoding the video sequence with our RQ-LVC, respectively. We evaluate with 4 quality levels in $\left\{10, 25, 40, 55 \right\}$ and report the average per-sequence rate deviation over these quality levels.

The accuracy $P^{RQNet}$ of RQNet in predicting the bitrates at predefined quality levels is defined as the average absolute difference between the predicted bitrate $R_{t}^{pred}$ and the actual bitrate $R_{t}^{enc}$ across $n$ frames:
\begin{equation}
\label{equ:acc_RQNet}
P^{RQNet}=\frac{\sum_{t=1}^{n}\frac{\left |R_{t}^{enc}-R_{t}^{pred} \right |}{R_{t}^{pred}}}{n}\times100\%,
\vspace{-0.4cm}
\end{equation}

To evaluate the rate-distortion performance, we calculate BD-rate savings~\cite{bdrate}, with the anchor adopting constant quality levels without rate control.
% For each video, we report the average per-frame bitrate deviation in percentage terms. Additionally, we present the complexity of the competing methods in terms of kMAC/pixel and model size. 

\textbf{Baseline Methods:} Our method is compared with the multi-pass encoding approach, which pre-encodes frames with four quality levels $\left\{10, 17, 43, 60\right\}$ and estimates model parameters using the least-squares algorithm. We adopt Li~\emph{et al.}'s update strategy~\cite{LiLiLambda} as our baseline. Following~\cite{LiLiLambda}, we use the same initial $\alpha_{t=1}$ and $\beta_{t=1}$ for each test sequence, obtained by averaging their respective values over the first frames of all sequences in both datasets. Based on Eq.~\eqref{equ:Q_real}, the quality level $Q^{real}_t$ is derived from the target bitrate $R^{target}_{t}$ and used to encode the current coding frame $X_t$ with MaskCRT, resulting in a bitrate of $R^{real}_t$:
\begin{equation}
\label{equ:Q_real}
Q^{real}_t = \alpha_{t} \times \operatorname{ln}(R^{target}_{t}) + \beta_{t}.
\end{equation}
% \begin{equation}
% \label{equ:R_real}
% R^{real_t} = Encode(X_{t}|Q^{real}_t),
% % \vspace*{-.1cm}
% \end{equation}
With $R^{real}_t$, the corresponding quality level $Q^{est}_t$ is estimated as:
\begin{equation}
\label{equ:Q_est}
Q^{est}_t = \alpha_{t} \times \operatorname{ln}(R^{real}_t) + \beta_{t}.
% \vspace*{-.1cm}
\end{equation}
By adopting the Adaptive Least Mean Square method, the model parameters $\alpha, \beta$ are updated iteratively by Eq.~\eqref{equ:alpha_new}  and Eq.~\eqref{equ:beta_new}, respectively:
\begin{align}
\label{equ:alpha_new}
\alpha_{t+1} &= \alpha_t + \mu \times (Q^{real}_t - Q^{est}_t) \times \ln(R^{real}_t), \\ 
\label{equ:beta_new}
\beta_{t+1} &= \beta_t + \eta \times (Q^{real}_t - Q^{est}_t),
\end{align}
where $\mu$ and $\eta$ are the learning rates, which are set to 0.01 in our experiments.
Besides, we adopt the R-Q model and updating mechanism proposed by Liao~\emph{et al.}~\cite{RC_ICASSP} as a competing method.
% Note that the first 2 P-frames after the I-frame are encoded using random quality levels, applied to all methods for fair comparison.

\subsection{Experimental Results}

% \textbf{Prediction Accuracy of RQNet and Its Complexity:}   Table~\ref{tab:complexity} compares its complexity with that of the base codec, MaskCRT. RQNet's kMAC/pixel is seen to be less than one-tenth of that of MaskCRT, and its model size is only a fraction of MaskCRT's.

%The prediction of the $\alpha$ and $\beta$ coefficients can be performed with just four points predicted by RQNet. However, the accuracy of this estimation may be limited. Therefore, additional (R,Q) points from previously coded frames are needed to enhance accuracy.
%presents a feasible solution with very low complexity to collect the set of points (R,Q), eliminating the need to perform the full frame encoding process at various quality levels.
% \input{chapters/tables/compare_weight}
\begin{figure*}[t!]
    \begin{center}
    % \vspace{-1.0em}
    \begin{subfigure}{0.31\linewidth}
        \centering
        \includegraphics[width=\linewidth]{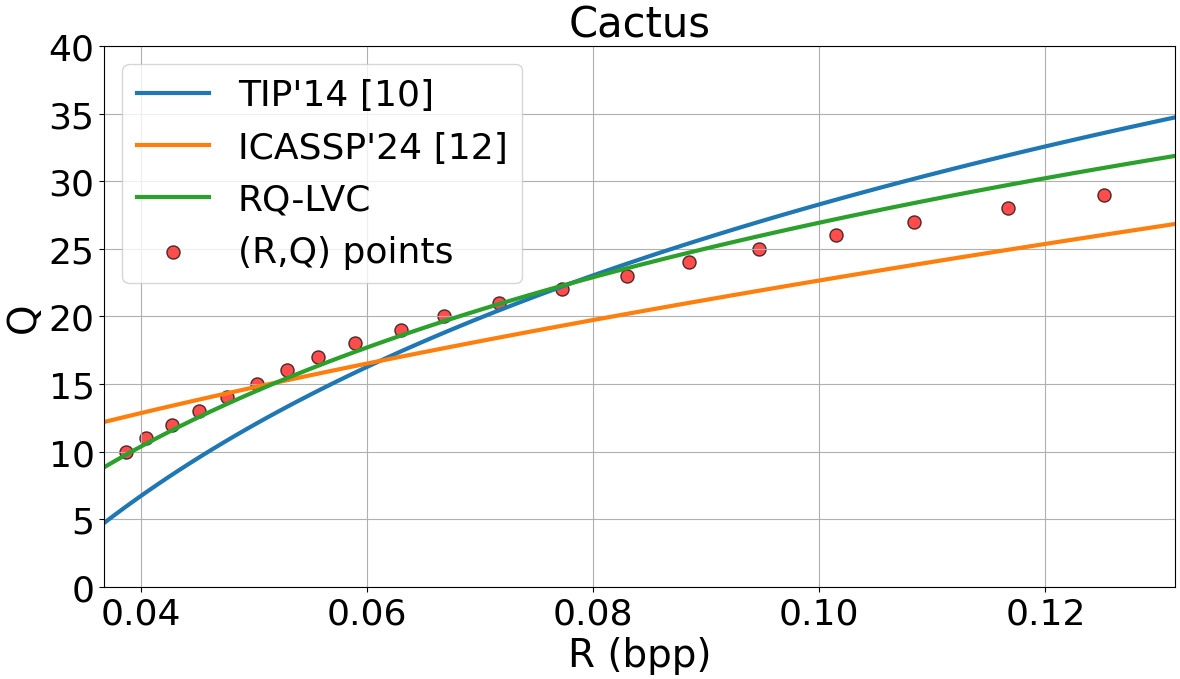}
    \end{subfigure}
    \begin{subfigure}{0.31\linewidth}
        \centering
        \includegraphics[width=\linewidth]{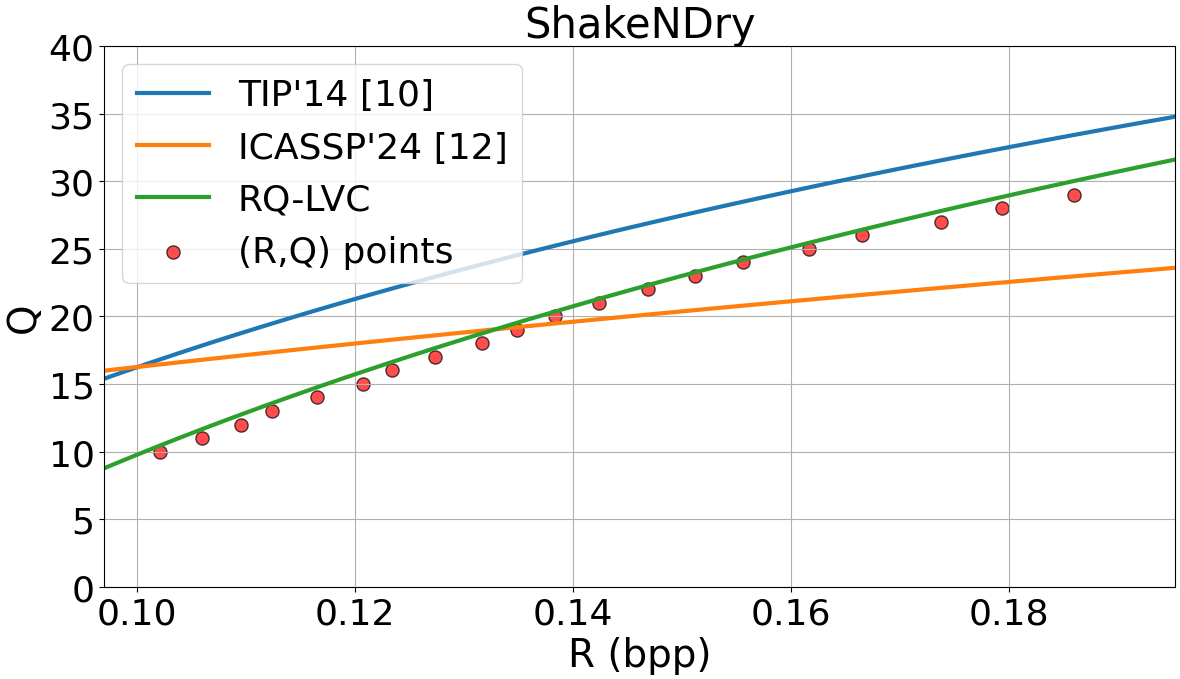}
    \end{subfigure}
    \begin{subfigure}{0.31\linewidth}
        \centering
        \includegraphics[width=\linewidth]{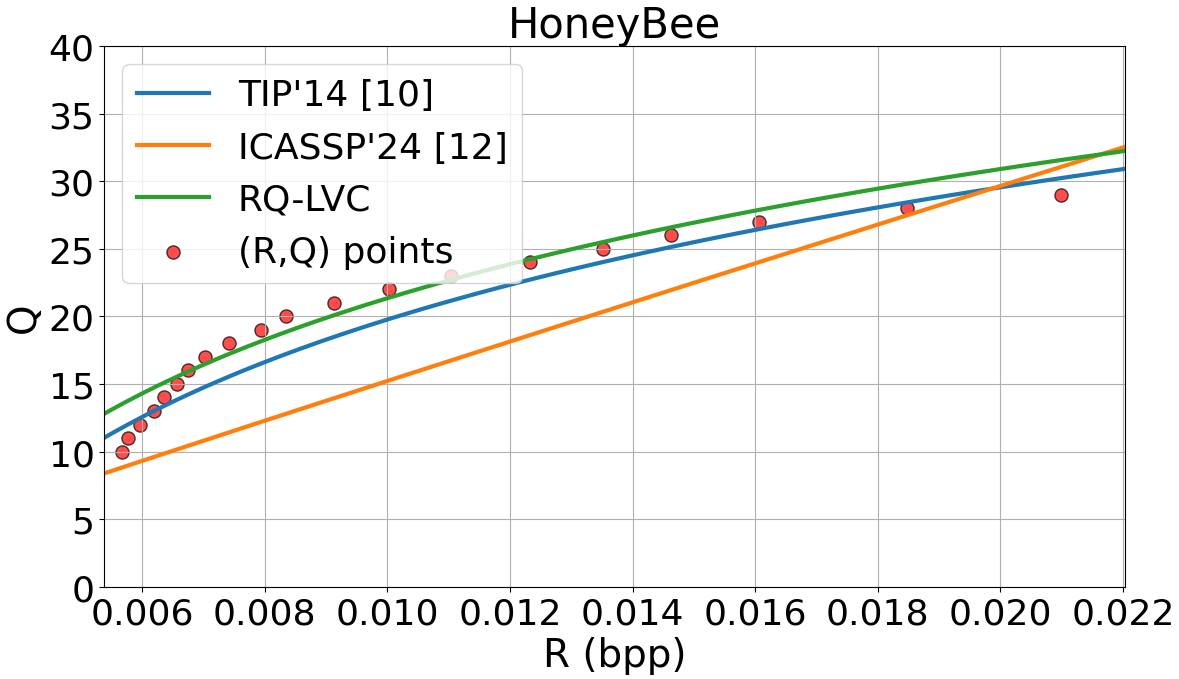}
    \end{subfigure}
    \vspace{-0.1em}
    \caption{The predicted R-Q functions for frame 16 versus the ground-truth (R,Q) points collected by multi-pass encoding.}
    \label{fig:curve_visualize}
    \end{center}
    \vspace{-1.5em}
% \vspace{-0.3em}
\end{figure*}
\begin{table*}[t]
\centering
\footnotesize
% \small
% \setlength{\tabcolsep}{3pt}
\caption{Comparison of the competing rate control schemes in terms of bitrate deviation $\Delta R^{RC}$ and BD-rate saving (\%). Notation: $\Delta R^{RC}$ / BD-rate saving (\%). 
% The best and second best of the one-pass methods are in Bold and Underlined, respectively.
}
% \fontsize{7.5}{9}\selectfont

% \fontsize{9}{10}\selectfont
\begin{tabular}{cccccccc}
\hline
\multirow{2}{*}{Dataset}                                                & \multirow{2}{*}{Sequence}                          & \multirow{2}{*}{Four-pass} & \multicolumn{5}{c}{One-pass}                                                                                                                                                                                                                                       \\ \cline{4-8} 
                                                                        &                           &                             & TIP'14~\cite{LiLiLambda} & ICASSP'24~\cite{RC_ICASSP} & RQ-LVC w/ RQNet Only & RQ-LVC w/o RQNet & RQ-LVC        \\ \hline
\multirow{8}{*}{UVG}                                                    & Beauty                    & 1.27 / 7.2                  & 2.15 / 0.4                                              & 3.47 / 2.5                                                  & 6.47 / 7.2                                                     & 1.70 / 1.2                                                 & 2.07 / -0.1   \\
                                                                        & Bosphorus                 & 2.42 / -5.8                 & 4.38 / -2.9                                             & 1.41 / -5.9                                                & 7.50 / -5.8                                                    & 0.82 / -2.7                                                & 0.46 / -6.1   \\
                                                                        & HoneyBee                  & 4.28 / -7.8                 & 6.63 / -3.2                                            & 34.09 / -3.3                                               & 2.88 / -4.8                                                    & 13.25 / -20.5                                              & 2.23 / -7.7   \\
                                                                        & Jockey                    & 1.69 / -2.9                 & 4.99 / -0.4                                             & 1.71 / 0.0                                                 & 4.18 / -2.0                                                    & 0.74 / -3.7                                                & 1.58 / -3.5   \\
                                                                        & ReadySteadyGo             & 1.80 / -4.3                 & 2.96 / -2.3                                             & 1.97 / -4.5                                                & 4.47 / -2.8                                                    & 1.15 / -3.0                                                & 0.84 / -4.8   \\
                                                                        & ShakeNDry                 & 0.19 / -2.6                 & 5.97 / -1.9                                             & 2.65 / -0.8                                                & 2.21 / -2.1                                                    & 1.31 / -3.4                                                & 0.88 / -2.6   \\
                                                                        & YachtRide                 & 0.42 / -2.3                 & 3.21 / -2.2                                             & 2.27 / -0.8                                                & 1.09 / -2.0                                                    & 1.38 / -2.2                                                & 1.04 / -2.3   \\
                                                                        & \textit{Average}          & \textit{1.72 / -2.6 }         & \textit{4.33 / -1.8}                                           & \textit{6.80 / -1.8}                                              & \textit{4.11 / -1.8}                                                  & \textit{2.90 / -4.9}                                                       & \textit{1.30 / -3.9} \\ \hline
\multirow{6}{*}{HEVC-B} & BasketballDrive           & 1.00 / -2.3                 & 2.24 / -2.2                                             & 3.65 / -2.1                                                & 1.77 / -2.7                                                    & 2.21 / -2.2                                                & 1.63 / -2.9   \\
                                                                        & BQTerrace                 & 1.64 / -2.1                 & 11.69 / 7.3                                             & 1.96 / -1.1                                                & 1.10 / -1.6                                                    & 3.58 / 0.2                                                 & 0.88 / -1.8   \\
                                                                        & Cactus                    & 1.42 / -4.0                 & 1.08 / -5.8                                             & 2.72 / -3.7                                                & 2.27 / -3.0                                                    & 1.88 / -3.7                                                & 0.57 / -4.1   \\
                                                                        & Kimono1                   & 0.20 / 0.5                  & 2.12 / -0.3                                             & 2.52 / 1.1                                                 & 2.60 / -0.4                                                    & 0.81 / -0.3                                                & 0.40 / -0.4   \\
                                                                        & ParkScene                 & 1.89 / -5.0                 & 0.96 / -8.1                                             & 2.53 / -4.6                                                & 2.04 / -3.2                                                    & 1.45 / -4.7                                                & 0.59 / -4.8   \\
                                                                        & \textit{Average}          & \textit{1.23 / -2.6}         & \textit{3.62 / -1.8}                                           & \textit{2.67 / -2.1}                                              & \textit{1.96 / -2.2}                                                  & \textit{1.99 / -2.1}                                              & \textit{0.81 / -2.8} \\ \hline
\end{tabular}
\label{tab:Rate_Control_96F_FullRC}
% \vspace{-0.5em}
\vspace{-0.4cm}
\end{table*}
% \usepackage{multirow}
\begin{table}[t!]
\centering
% \normalsize
\scriptsize
\setlength{\tabcolsep}{5pt}
\caption{Prediction accuracy $P^{RQNet}$ of RQNet in estimating bitrates at predefined quality levels.}
\begin{tabular}{>{\centering\arraybackslash}p{1.0cm} >{\centering\arraybackslash}p{1.6cm} >{\centering\arraybackslash}p{0.6cm}>{\centering\arraybackslash}p{0.6cm}>{\centering\arraybackslash}p{0.6cm}>{\centering\arraybackslash}p{0.6cm}>{\centering\arraybackslash}p{0.8cm}}
\hline
Dataset                 & Sequence         & Q=10           & Q=27           & Q=43           & Q=60           & Average        \\ \hline
\multirow{8}{*}{UVG}    & Beauty           & 7.01           & 4.50           & 24.01          & 24.12          & 14.91          \\
                        & Bosphorus        & 41.21          & 28.77          & 27.54          & 23.29          & 30.20          \\
                        & HoneyBee         & 11.63          & 17.44          & 44.59          & 38.90          & 28.14          \\
                        & Jockey           & 20.75          & 12.00          & 12.93          & 17.20          & 15.72          \\
                        & ReadySteadyGo    & 11.87          & 13.70          & 20.66          & 24.94          & 17.79          \\
                        & ShakeNDry        & 12.99          & 12.72          & 15.82          & 15.45          & 14.25          \\
                        & YachtRide        & 7.40           & 10.86          & 8.27           & 12.30          & 9.71           \\
                        & \textit{Average} & \textit{16.12} & \textit{14.28} & \textit{21.98} & \textit{22.31} & \textit{16.87} \\ \hline
\multirow{6}{*}{HEVC-B} & Kimono1          & 5.99           & 7.94           & 6.26           & 19.17          & 9.84           \\
                        & BQTerrace        & 23.11          & 12.77          & 23.82          & 25.29          & 21.25          \\
                        & Cactus           & 18.30          & 12.90          & 20.34          & 21.12          & 18.16          \\
                        & BasketballDrive  & 8.50           & 6.82           & 13.90          & 5.57           & 8.70           \\
                        & ParkScene        & 21.04          & 17.03          & 22.64          & 24.84          & 21.39          \\
                        & \textit{Average} & \textit{15.39} & \textit{11.49} & \textit{17.39} & \textit{19.20} & \textit{15.87} \\ \hline
\end{tabular}
\label{tab:RQNet_acc}
\vspace{-0.5cm}
\end{table}

\textbf{Rate Control Performance of RQ-LVC:} Table~\ref{tab:Rate_Control_96F_FullRC} evaluates the rate control performance of our RQ-LVC framework. As shown in the table, RQ-LVC consistently achieves significantly smaller bitrate deviations than the baseline methods~\cite{LiLiLambda,RC_ICASSP}. This improvement is attributed to the fact that RQ-LVC leverages the (R,Q) points output by RQNet, along with those from previously coded frames, to estimate model parameters $\alpha,\beta$. In contrast, the approaches presented by Li~\emph{et al.}~\cite{LiLiLambda} and Liao~\emph{et al.}~\cite{RC_ICASSP} rely exclusively on the (R,Q) points from the past frames. Fig.~\ref{fig:curve_visualize} visualizes the  R-Q functions with the predicted $\alpha,\beta$ for frame 16 across various video sequences, alongside the actual (R,Q) points obtained by encoding the video with various quality levels. The R-Q function predicted by our RQ-LVC closely aligns with the actual (R,Q) points, whereas the R-Q functions predicted by Li~\emph{et al.}~\cite{LiLiLambda} exhibit significant deviations from these ground truths. In some sequences, such as Bosphorus, HoneyBee, Jockey, and ReadySteadyGo, our proposed RQ-LVC even outperforms the time-consuming multi-pass encoding strategy. 
In addition, the application of hierarchical quality-level patterns allows efficient compression. Our RQ-LVC ensures that actual bitrates closely match the allocated target bitrates for every frame, thereby achieving consistently high compression performance across all test sequences.
Besides, the additional overhead is minimal compared to MaskCRT, the base codec. Specifically, the additional complexity introduced by RQNet amounts to an 8\% increase in encoding multiply-accumulate operations per pixel (kMAC/pixel) and a 14\% increase in encoding time, while the RQNet model size represents only a fraction of MaskCRT's.
% As presented in Table~\ref{tab:complexity}, the additional overhead introduced by RQNet is minimal compared to MaskCRT, the base codec, both in terms of kMAC/pixel and model size. These attributes make RQ-LVC a practical solution for rate control systems, particularly in scenarios where multi-pass methods are infeasible.

%using the (R,Q) points predicted by RQNet in combination with those gathered from previous frames to estimate $\alpha,\beta$ yields effective quality level predictions. 

% It is also shown that using the weighted least-squares method to predict the parameters is more effective than the conventional method without using weight. This is explained by the fact that frames closer to the current frame are more closely related, so a larger weight should be used for these frames and vice versa. 
% However, in most cases, our proposed method does not perform as well as the approach by Li~\emph{et al.}~\cite{LiLiLambda}, as in the initial frames, our data come primarily from RQNet, whereas Li's method utilizes actual (R,Q) points to initialize $\alpha$ and $\beta$.

\textbf{Model parameter estimation with RQNet Only:} We evaluate the rate control performance of RQNet by estimating model parameters solely from RQNet-predicted (R,Q) points, in a way similar to~\cite{rate_fitting}, which uses neural networks to predict (R,$\lambda$) and (D,$\lambda$) data points to estimate model parameters for rate control. From Table~\ref{tab:Rate_Control_96F_FullRC}, it is evident that relying exclusively on the (R,Q) points from RQNet is ineffective. To further analyze the performance of RQNet, Table~\ref{tab:RQNet_acc} presents RQNet's prediction accuracy. RQNet achieves average bitrate deviations of 16.87\% and 15.87\% on UVG and HEVC Class B datasets, respectively. The deviations are relatively larger at higher bitrates, indicating that the predicted (R,Q) points are less reliable at higher rates, making it beneficial to also use the (R,Q) points from previously coded frames.

\textbf{Model parameter estimation without RQNet:} The results in Table~\ref{tab:Rate_Control_96F_FullRC} also demonstrate that model parameter estimation using only previously coded (R,Q) points can still yield effective results. However, this approach has two significant limitations: (1) it requires a substantial number of historical (R,Q) points, and (2) it struggles when the target bitrate range maps to an extremely narrow QP range, as observed in the HoneyBee sequence results.
\begin{figure}[t]
\centering
\includegraphics[width=\linewidth]{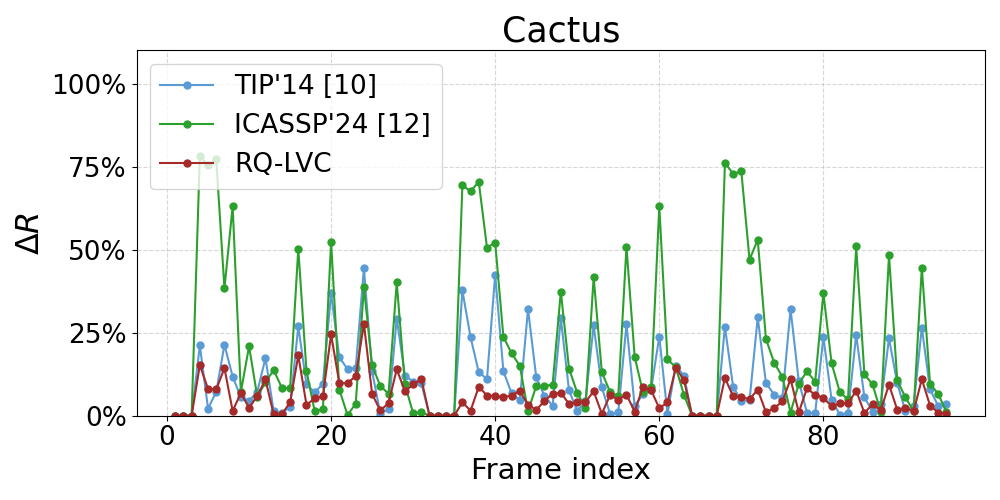}
    %\vspace{-3em}
\caption{Comparison of per-frame rate deviation $\Delta R$ between our RQ-LVC and competing methods for Cactus sequence.}
\label{fig:per_deviation}
\vspace{-0.5cm}
\end{figure}

\textbf{Per-frame rate deviation:} Maintaining precise rate control is critical, as deviations impact both current and subsequent frame quality. Exceeding the target bitrate constrains the bitrate budget for remaining frames, while under-encoding compromises the quality of the current frame. As demonstrated in Fig.~\ref{fig:per_deviation}, RQ-LVC maintains this accuracy even for the first initial frames within a GOP, where historical coding data is limited. This is acheived by utilizing predictions from RQNet. In contrast, conventional methods~\cite{LiLiLambda}~\cite{RC_ICASSP}
% without neural network support
% (e.g., Li~\emph{et al.}~\cite{LiLiLambda}, Liao~\emph{et al.}~\cite{RC_ICASSP}) 
frequently struggle in such scenarios.

\section{Conclusion}
This paper proposes a rate-quality model for learned video coding and an adaptive strategy for on-the-fly parameter updates. We introduce RQNet, a lightweight neural network that predicts encoding bits at predefined quality levels, removing the need for multiple encoding passes. By combining RQNet's predictions with the encoding results from previously coded frames, we explore the least-squares method to estimate the model parameters for each frame. Experimental results show our method achieves significantly lower bitrate deviations than the baseline method, with minimal extra computational overhead. For future work, we will improve RQNet's effectiveness, weight (R,Q) data point contributions to improve parameter estimation, and developing neural approaches for bitrate allocation to improve compression performance.

\printbibliography

@article{hevc,
  title={{O}verview of the {H}igh {E}fficiency {V}ideo {C}oding ({HEVC}) {S}tandard},
  author={Sullivan, Gary J and Ohm, Jens-Rainer and Han, Woo-Jin and Wiegand, Thomas},
  journal={IEEE Transactions on Circuits and Systems for Video Technology},
  volume={22},
  number={12},
  pages={1649--1668},
  year={2012},
  publisher={IEEE}
}

@article{overview_vvc,
  title={{O}verview of the {V}ersatile {V}ideo {C}oding ({VVC}) {S}tandard and {I}ts {A}pplications},
  author={Bross, Benjamin and Wang, Ye-Kui and Ye, Yan and Liu, Shan and Chen, Jianle and Sullivan, Gary J and Ohm, Jens-Rainer},
  journal={IEEE Transactions on Circuits and Systems for Video Technology},
  volume={31},
  number={10},
  pages={3736--3764},
  year={2021},
  publisher={IEEE}
}

@inproceedings{uvg,
  title={{UVG} {D}ataset: 50/120fps 4{K} {S}equences for {V}ideo {C}odec {A}nalysis and {D}evelopment},
  author={Mercat, Alexandre and Viitanen, Marko and Vanne, Jarno},
  booktitle={Proceedings of the 11th ACM Multimedia Systems Conference},
  pages={297--302},
  year={2020}
}

@article{hevcctc,
  title={Common test conditions and software reference configurations},
  author={Bossen, Frank and others},
  journal={JCTVC-L1100},
  year={2013}
}

@article{vimeo,
  title={{V}ideo {E}nhancement with {T}ask-{O}riented {F}low},
  author={Xue, Tianfan and Chen, Baian and Wu, Jiajun and Wei, Donglai and Freeman, William T},
  journal={International Journal of Computer Vision},
  volume={127},
  number={8},
  pages={1106--1125},
  year={2019},
  publisher={Springer}
}

@inproceedings{canfvc,
  title={{CANF-VC}: {C}onditional {A}ugmented {N}ormalizing {F}lows for {V}ideo {C}ompression},
  author={Ho, Yung-Han and Chang, Chih-Peng and Chen, Peng-Yu and Gnutti, Alessandro and Peng, Wen-Hsiao},
  booktitle={2022 European Conference on Computer Vision (ECCV)},
  pages={207--223},
  year={2022}
}

@ARTICLE{h264,  author={Wiegand, T. and Sullivan, G.J. and Bjontegaard, G. and Luthra, A.},  journal={IEEE Transactions on Circuits and Systems for Video Technology},   title={{O}verview of the {H.264/AVC} {V}ideo {C}oding {S}tandard},   year={2003},  volume={13},  number={7},  pages={560-576}}

@InProceedings{dcvcdc,
  title={{N}eural {V}ideo {C}ompression with {D}iverse {C}ontexts},
  author={Li, Jiahao and Li, Bin and Lu, Yan},
  booktitle={2023 IEEE/CVF Conference on Computer Vision and Pattern Recognition (CVPR)},
  pages={22616-22626},
  year={2023}
}

@article{bdrate,
  title={{W}orking {P}ractices {U}sing {O}bjective {M}etrics for {E}valuation of {V}ideo {C}oding {E}fficiency {E}xperiments},
  journal={Standard ISO/IEC TR 23002-8, ISO/IEC JTC 1, Jul},
  year={2020}
}

@inproceedings{dcvcfm,
  title={{N}eural {V}ideo {C}ompression with {F}eature {M}odulation},
  author={Li, Jiahao and Li, Bin and Lu, Yan},
  booktitle={2024 IEEE/CVF Conference on Computer Vision and Pattern Recognition (CVPR)},
  pages={26099-26108},
  year={2024}
}

@Article{maskcrt,
  title= "{M}ask{CRT}: {M}asked {C}onditional {R}esidual {T}ransformer for {L}earned {V}ideo {C}ompression",
  author= "Chen, Yi-Hsin and Xie, Hong-Sheng and Chen, Cheng-Wei and Gao, Zong-Lin and Peng, Wen-Hsiao and Benjak, Martin and Ostermann, Jörn",
  journal={IEEE Transactions on Circuits and Systems for Video Technology},
  year={2024},
  volume={34},
  number={11},
  pages={11980-11992},
}

@INPROCEEDINGS{rd_modeling_lic,
  author={Jia, Chuanmin and Ge, Ziqing and Wang, Shanshe and Ma, Siwei and Gao, Wen},
  booktitle={2022 Data Compression Conference (DCC)}, 
  title={{R}ate {D}istortion {C}haracteristic {M}odeling for {N}eural {I}mage {C}ompression}, 
  year={2022},
  volume={},
  number={},
  pages={202-211},
}

@article{LambdaDomainRC_LIC,
  title={Lambda-{D}omain {R}ate {C}ontrol for {N}eural {I}mage {C}ompression},
  author={Naifu Xue and Yuan Zhang},
  journal={Proceedings of the 5th ACM International Conference on Multimedia in Asia},
  year={2023},
}

@article{LiLiLambda,
  title={$\lambda$  {D}omain {R}ate {C}ontrol {A}lgorithm for {H}igh {E}fficiency {V}ideo {C}oding},
  author={Bin Li and Houqiang Li and Li Li and Jinlei Zhang},
  journal={IEEE Transactions on Image Processing},
  year={2014},
  volume={23},
  pages={3841-3854},
}

@INPROCEEDINGS{RC_ICASSP,
  author={Liao, Shuhong and Jia, Chuanmin and Fan, Hongfei and Yan, Jingwen and Ma, Siwei},
  booktitle={ICASSP 2024 - 2024 IEEE International Conference on Acoustics, Speech and Signal Processing (ICASSP)}, 
  title={{R}ate-{Q}uality {B}ased {R}ate {C}ontrol {M}odel for {N}eural {V}ideo {C}ompression}, 
  year={2024},
  volume={},
  number={},
  pages={4215-4219},
}

@Article{SparseToDense,
    author = {Chen, Jiancong and Wang, Meng and Zhang, Pingping and Wang, Shurun and Wang, Shiqi},
    title = {{S}parse-to-{D}ense: {H}igh {E}fficiency {R}ate {C}ontrol for {E}nd-to-{E}nd {S}cale-{A}daptive {V}ideo {C}oding},
    journal = "IEEE Transactions on Circuits and Systems for Video Technology",
    year = {2024},
    issue_date = {May 2024},
    publisher = {IEEE Press},
    volume = {34},
    number = {5},
}

@inproceedings{neural_rate_control,
  title={{N}eural {R}ate {C}ontrol for {L}earned {V}ideo {C}ompression},
  author={Zhang, Yiwei and Lu, Guo and Chen, Yunuo and Wang, Shen and Shi, Yibo and Wang, Jing and Song, Li},
  booktitle={The Twelfth International Conference on Learning Representations},
  year={2023}
}

@book{bayesian,
  title={Bayesian Data Analysis},
  author={Gelman, Andrew and Carlin, John B and Stern, Hal S and Dunson, David B and Vehtari, Aki and Rubin, Donald B},
  edition={3rd},
  year={2013},
  publisher={Chapman and Hall/CRC},
  doi={10.1201/b16018}
}

@inproceedings{rate_control_tencent,
  author={Li, Yanghao and Chen, Xinyao and Li, Jisheng and Wen, Jiangtao and Han, Yuxing and Liu, Shan and Xu, Xiaozhong},
  booktitle={2022 IEEE International Conference on Acoustics, Speech and Signal Processing (ICASSP)}, 
  title={{R}ate {C}ontrol for {L}earned {V}ideo {C}ompression}, 
  year={2022},
  volume={},
  number={},
  pages={2829-2833}
}

@INPROCEEDINGS{rate_fitting,
  author={Gu, Bowen and Chen, Hao and Lu, Ming and Yao, Jie and Ma, Zhan},
  booktitle={2025 Data Compression Conference (DCC)}, 
  title={{A}daptive {R}ate {C}ontrol for {D}eep {V}ideo {C}ompression with {R}ate-{D}istortion {P}rediction}, 
  year={2025},
  volume={},
  number={},
  pages={33-42},
}

@inproceedings{depthCNN,
  author={Chollet, François},
  booktitle={2017 IEEE Conference on Computer Vision and Pattern Recognition (CVPR)}, 
  title={Xception: {D}eep {L}earning with {D}epthwise {S}eparable {C}onvolutions}, 
  year={2017},
  volume={},
  number={},
  pages={1800-1807},
}

\end{document}